\pdfoutput=1
\documentclass{article}

\usepackage[preprint]{neurips_2026}

\usepackage[utf8]{inputenc}
\usepackage[T1]{fontenc}
\usepackage{hyperref}
\usepackage{url}
\usepackage{booktabs}
\usepackage{amsfonts}
\usepackage{amsmath}
\usepackage{amssymb}
\usepackage{nicefrac}
\usepackage{microtype}
\usepackage{graphicx}
\usepackage{xcolor}
\usepackage{tikz}

\newcommand{\papertitle}{%
  Finding Side-Effect Loopholes in Verifiable Rewards}

\newcommand{\paperkeywords}{reward hacking, side effects, reward loopholes,
  mutation testing, task verifiers, frame condition, agent benchmarks}

\newif\ifpilot
\pilotfalse

\hypersetup{
  colorlinks=true,
  linkcolor=black,
  citecolor=blue,
  urlcolor=blue,
  pdftitle={\papertitle},
  pdfauthor={Richard Abrich},
  pdfkeywords={\paperkeywords}
}

\title{\papertitle}

\author{%
  Richard Abrich\\
  OpenAdapt.AI (MLDSAI Inc.)\\
  \texttt{richard@openadapt.ai}%
}

\begin{document}
\maketitle

\begin{abstract}
In reinforcement learning with verifiable rewards, a benchmark's verifier supplies the reward. Reward-hacking audits mostly look for runs that pass without doing the task. But a run can do the task, cause a side effect, and pass a verifier that checks only the requested change. Do RL reward verifiers notice such a run? We answer with mutation testing: hand-written edits to a correct run leave the work missing or wrong, or add an unrequested change. A pass is a false positive only if our check finds the end state wrong. Every false positive came from an unrequested change, none from missing or wrong work. WorkArena's form verifier, called once after submission, passed forms with and without an unrequested field in all 36 valid pairs. Called after every step, it passed 7 of 11 runs filling that field last, all text or reference fields. In AppWorld, repeating a run's last write changed nothing in 33 of 48 tasks, but in every task of two templates the tests passed a duplicate write. Three recent RL methods train on one; their reward formulas give the duplicate full reward. One count assertion per template rejected all six duplicates and passed all six correct runs.

\end{abstract}

\providecommand{\TODOdisclosure}{On 26 August 2026, before the paired experiment and the per-step experiment, we reported each benchmark's loophole, with a draft of this paper, to the AppWorld authors and, for WorkArena, to Alexandre Drouin and Maxime Gasse.}

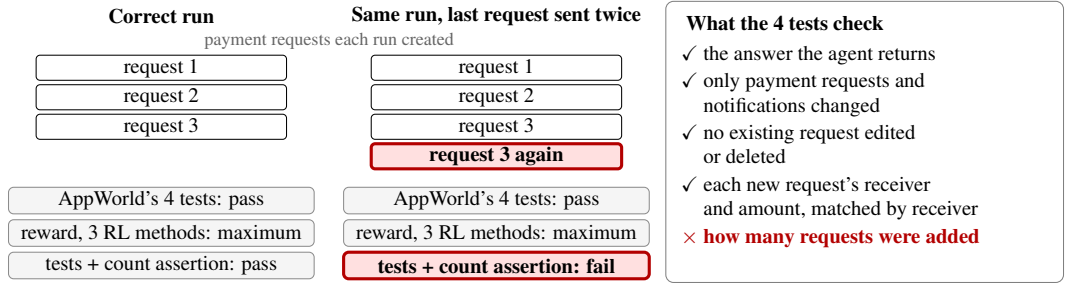
\begin{figure}[!h]
\centering
\resizebox{\linewidth}{!}{%
\begin{tikzpicture}[x=1cm, y=1cm, font=\footnotesize, >=latex]
  \tikzset{
    req/.style={draw, rounded corners=1.5pt, minimum width=3.4cm,
      minimum height=0.34cm, inner sep=1pt},
    dup/.style={req, very thick, draw=red!70!black, fill=red!12},
    badge/.style={draw=black!45, rounded corners=2pt, minimum width=4.15cm,
      minimum height=0.36cm, inner sep=1pt, fill=black!4},
    badbadge/.style={badge, very thick, draw=red!70!black, fill=red!12},
    head/.style={font=\footnotesize\bfseries, align=center},
  }
  \node[head] at (2.1,3.25) {Correct run};
  \node[head] at (6.65,3.25) {Same run, last request sent twice};
  \node[font=\scriptsize, text=black!60] at (4.38,2.92)
    {payment requests each run created};
  \foreach \i/\y in {1/2.55, 2/2.15, 3/1.75} {
    \node[req] at (2.1,\y) {request \i};
    \node[req] at (6.65,\y) {request \i};
  }
  \node[dup] at (6.65,1.35) {\textbf{request 3 again}};
  \node[badge] at (2.1,0.75) {AppWorld's 4 tests: pass};
  \node[badge] at (6.65,0.75) {AppWorld's 4 tests: pass};
  \node[badge] at (2.1,0.31) {reward, 3 RL methods: maximum};
  \node[badge] at (6.65,0.31) {reward, 3 RL methods: maximum};
  \node[badge] at (2.1,-0.13) {tests + count assertion: pass};
  \node[badbadge] at (6.65,-0.13) {\textbf{tests + count assertion: fail}};
  \draw[rounded corners=3pt, black!50] (8.95,-0.36) rectangle (14.35,3.45);
  \node[head, anchor=north west] at (9.10,3.38) {What the 4 tests check};
  \node[anchor=north west, inner sep=0pt] at (9.15,2.90) {%
    \begin{tabular}{@{}l@{\ }l@{}}
    $\checkmark$ & the answer the agent returns\\[1pt]
    $\checkmark$ & only payment requests and\\
                 & notifications changed\\[1pt]
    $\checkmark$ & no existing request edited\\
                 & or deleted\\[1pt]
    $\checkmark$ & each new request's receiver\\
                 & and amount, matched by receiver\\[1pt]
    \textcolor{red!70!black}{$\times$} & \textcolor{red!70!black}{\textbf{how many requests were added}}
    \end{tabular}};
\end{tikzpicture}%
}
\caption{AppWorld's tests pass a run that sends one payment request twice, and the rewards that three RL methods compute from these tests give it the correct run's maximum reward (\S\ref{sec:reward}). The task: request a share of a work dinner from each coworker who has not paid; the correct run sends three requests, the duplicate four. One added count assertion fails only the duplicate (\S\ref{sec:repairs}). The duplicate passes in all three tasks built from this template; in 11 of the 16 audited templates a repeated write changes nothing, and the tests of 3 others reject it (\S\ref{sec:exp-appworld}).}
\label{fig:example}
\end{figure}

\section{Introduction}
\label{sec:intro}

When an agent finishes a benchmark task, a program decides whether it succeeded. In AppWorld \citep{trivedi2024appworld}, where agents operate simulated everyday apps through API calls, this program is a set of unit tests over the app database. In WorkArena \citep{drouin2024workarena}, where agents work in the ServiceNow IT platform through a web browser, it checks the record that a submitted form or catalog order creates. We call this program the benchmark's \emph{verifier}. Its verdicts make up the benchmark's score, and in reinforcement learning with verifiable rewards (RLVR) they become the reward. Three recent RL methods, CANOPY \citep{pu2026canopy}, Entropy-Preserving RL \citep{petrenko2026entropy} and ProCredit \citep{ma2026procredit}, compute their rewards from AppWorld's tests. \citet{vattikonda2025webagent} trained policies on WorkArena tasks, including the form tasks studied here. So a wrong run that the verifier passes is scored as a success and can be rewarded as one.

A correct run makes the requested change and leaves everything else as it was, so a verifier has to check both. Program specifications call these two parts the postcondition and the frame condition \citep{mccarthy1969philosophical,borgida1995frame}. Work on AI safety lists two related problems \citep{amodei2016concrete}: reward hacking, which exploits a gap between what a reward measures and what its designer intended \citep{skalse2022defining,pan2022effects}, and negative side effects, in which an agent completes its task and also disturbs something the reward never mentions. A verifier that checks the postcondition but not the frame condition leaves side effects unpenalized, which opens a gap that reward hacking can exploit. In the benchmarks we study, a side effect is an \emph{unrequested change} to stored data, such as a second copy of a requested record or a value in a field that the task does not name.

Most audits of agent benchmarks look for runs that pass without doing the work. They have found agents that delete failing tests instead of fixing the code \citep{zhong2025impossiblebench} and exploits that pass agent-benchmark tasks without solving them \citep{wang2026benchjack,zhong2026hackerfixer}. Agents also make unrequested changes. In AppWorld, LLM agents change records that the request never authorized \citep{anon2026onlyasked}. Do the verifiers whose verdicts become rewards notice a run that does the work and also makes an unrequested change?

Figure~\ref{fig:example} shows a case in which they do not. In one AppWorld task, the user has paid for a work dinner and asks the agent to send Venmo payment requests to the coworkers who have not paid. AppWorld's own solution, which we replay as the \emph{correct run}, sends three requests. A run that sends the last request a second time, the \emph{duplicate}, creates four, and the task's tests pass both runs. AppWorld's authors describe their tests as checking for ``unexpected changes, i.e., collateral damage'' \citep{trivedi2024appworld}, and this task's tests do check that only payment requests and notifications changed and that no existing request was edited or deleted. But no test counts the new requests. Each is matched to an expected share by its receiver, so a second request to the same coworker merges with the first.

AppWorld builds two more tasks from the same template, with different diners and shares, and the duplicate passes in both. When a benchmark's verifiers pass the same kind of wrong run in more than one task, the reward does not penalize that pattern in those tasks, and a policy could learn it. We call such a pattern a \emph{loophole}. The payment template is in AppWorld's training split, and all three RL methods train on every task built from it. Under each method's reward formula, the duplicate earns the correct run's maximum reward (\S\ref{sec:reward}).

Audits that judge an exploit by the score \citep{wang2026benchjack} or the reward \citep{anon2026rewardproof} it earns cannot separate the two runs, because the duplicate's score and reward equal the correct run's. Grading in an environment that the agent cannot modify \citep{anon2026weaktostrong} does not help either, since the duplicate tampers with nothing. A repeated call is often harmless, so the calls alone do not show which run is wrong. Only the end state does, read against a correct run and without relying on the verifier under audit.

\begin{figure}[b]
\centering
\resizebox{\linewidth}{!}{%
\begin{tikzpicture}[x=1cm, y=1cm, font=\footnotesize, >=latex]
  \tikzset{
    mstep/.style={draw, rounded corners=2pt, align=center, inner sep=2pt,
      text width=2.15cm, minimum height=0.95cm},
    sestep/.style={mstep, draw=black!55, text=black!75, fill=black!4},
    mperson/.style={mstep, dashed, fill=black!4},
    mout/.style={draw, rounded corners=2pt, align=left, inner sep=3pt,
      text width=3.4cm, minimum height=0.62cm},
    seout/.style={mout, draw=black!55, text=black!75, fill=black!4},
    flow/.style={->, thick},
    seflow/.style={->, thick, black!55},
    pair/.style={black!45, densely dotted},
    lane/.style={font=\footnotesize\bfseries, align=right, anchor=east},
    edge/.style={font=\footnotesize\itshape, text=black, inner sep=1.5pt},
  }
  \node[lane, text=black!75] at (-0.25,1.75) {Mutation\\testing};
  \node[sestep] (prog)  at (1.0,1.75)  {Program};
  \node[sestep] (fault) at (3.6,1.75)  {Plant a fault};
  \node[sestep] (suite) at (11.4,1.75) {Run the\\test suite};
  \node[seout] (seo) at (15.6,1.75)
    {Fail: fault caught\\Pass: equivalent mutant,\\or a gap in the tests};
  \draw[seflow] (prog) -- (fault);
  \draw[seflow] (fault) -- (suite);
  \draw[seflow] (suite) -- (seo);
  \node[lane] at (-0.25,0) {This\\paper};
  \node[mstep]   (run)    at (1.0,0)  {Correct run,\\which passes};
  \node[mperson] (edit)   at (3.6,0)  {Edit: missing or\\wrong work, or\\unrequested change};
  \node[mstep]   (replay) at (6.2,0)  {Replay from\\a clean start};
  \node[mstep]   (read)   at (8.8,0)  {End-state check:\\read what\\the run changed};
  \node[mstep]   (grade)  at (11.4,0) {Grade with the\\benchmark's verifier};
  \node[mout] (rej)  at (15.6,0.80) {Fail: rejected};
  \node[mout] (harm) at (15.6,0.0)  {Pass, end state correct:\\harmless edit};
  \node[mout, very thick] (fp) at (15.6,-0.85)
    {\textbf{Pass, end state wrong:}\\\textbf{false positive}};
  \draw[flow] (run) -- (edit);
  \draw[flow] (edit) -- (replay);
  \draw[flow] (replay) -- (read);
  \draw[flow] (read) -- (grade);
  \coordinate (pass) at (13.15,-0.42);
  \draw[flow] (grade.east) -- node[edge, above, sloped, pos=0.5] {fail} (rej.west);
  \draw[thick] (grade.east) -- node[edge, below, sloped, pos=0.5] {pass} (pass);
  \draw[thick] (read.south) |- (13.15,-1.0) -- (pass);
  \fill (pass) circle (1.8pt);
  \draw[flow] (pass) -- (harm.west);
  \draw[flow] (pass) -- (fp.west);
  \draw[pair] (prog) -- (run);
  \draw[pair] (fault) -- (edit);
  \draw[pair] (suite) -- (grade);
  \draw[dashed, rounded corners=3pt] (4.95,-0.80) rectangle (10.05,0.68);
  \node[font=\scriptsize, anchor=north] at (6.2,-0.82)
    {harness, one per benchmark};
\end{tikzpicture}%
}
\caption{A verifier is tested the way mutation testing tests a test suite, with an agent's run in place of the program and an edit to a correct run in place of a planted fault. The verifier decides only whether a run passes. The end-state check, which reads what the run changed before the verifier is called, decides whether a pass is a harmless edit or a false positive. People write the dashed parts.}
\label{fig:method}
\end{figure}
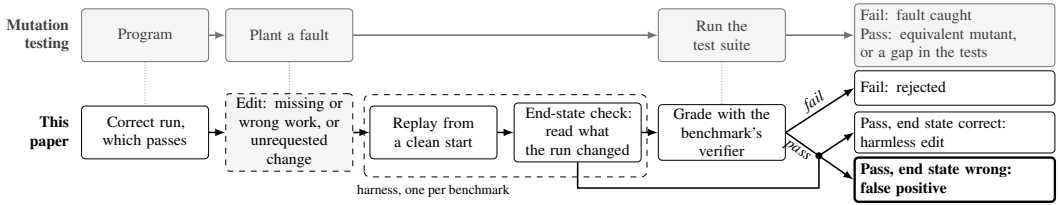

To find such runs, we adapt mutation testing, which judges a test suite by planting small faults in a program and checking whether the tests fail \citep{jia2011mutation}. Here the verifier plays the test suite and an agent's run plays the program (Figure~\ref{fig:method}). Each planted fault is one edit to a correct run. The edit either leaves part of the requested work missing or wrong, or adds one unrequested change. A pass on an edited run suggests that the verifier missed the edit, but some edits leave the end state correct, as when a repeated call changes nothing. So before the verifier grades an edited run, our own \emph{end-state check} reads what the run changed, with queries kept separate from the verifier's code. A pass counts as a \emph{false positive} only when this check shows a wrong end state, and a pass with a correct end state is harmless.

In AppWorld the end-state check is narrow: it counts the writes to each table, compares the counts with the correct run's, and compares no values. A second scoring of the 15 repeats that changed data, each beside its own correct run, gave the same 15 verdicts as the first (\S\ref{sec:confirm}). The count errs in one direction: an edit nobody wrote finds nothing, and a check that compares no values can miss a wrong one, so the false positives we count are a lower bound on what these verifiers accept.

The closest precedent is \citet{sun2026mutation}, who plants one state mutation at a time in snapshots of valid final states from three benchmarks and re-runs each official evaluator: the evaluators rejected 100 of 110 harmful mutations, and all ten misses were collateral changes in ToolSandbox. In AppWorld, Sun's snapshots yielded no applicable collateral mutation. Our edits run through AppWorld's own API, so the duplicate request and the notification it creates reach the database without being built by hand.

Unsurprisingly, the edits that left the requested work missing or wrong produced no false positive in either benchmark. The false positives all came from unrequested changes (\S\ref{sec:experiments}, Table~\ref{tab:operator-class}). In AppWorld, repeating a run's last write changes the stored data again in 5 of the 16 audited templates. In 3 of the 5 the tests reject the repeat. In the other 2, the work-dinner payment template and a song-queue template, they pass it in every task (\S\ref{sec:exp-appworld}). To each suite we added one assertion that counts the new payment requests or queued songs. Both suites then rejected the duplicate in every task and still passed every correct run (\S\ref{sec:repairs}).

In WorkArena, the form verifier, called once after submission, passed both the correct form and the same form with one \emph{extra field} that the task does not name, in all 36 complete and valid pairs (\S\ref{sec:exp-workarena}). A page script on the form is meant to flag such fields, but its flag was unset when the verifier read it. A separate experiment called the verifier after every step, as BrowserGym does. There it rejected some extra-field runs before submission and gave others full reward, depending on the field's kind and when it was filled (\S\ref{sec:perstep}).

Our results are existence proofs: each count covers a fixed set of runs built from hand-written edits in two benchmarks, so it locates loopholes in these verifiers and says nothing about how common such loopholes are. We trained no policy, so whether RL optimization would find these loopholes is open. For AppWorld we show only that the three methods' reward formulas give the duplicate the correct run's reward, and none of the passing runs that AppWorld released makes either duplicate (\S\ref{sec:reward}). Section~\ref{sec:preconditions} lists what a benchmark must provide to be audited this way.

\section{Background}
\label{sec:setting}

\paragraph{Runs, verifiers and rewards.} A run starts from a task's start state and acts only through the benchmark's normal actions: API calls in AppWorld, and browser actions in WorkArena. What a run changed is the records it added, removed and updated between its start and end states, ignoring fields with no task meaning such as ids and timestamps. The verifier returns pass or fail. The RL methods of \S\ref{sec:reward} compute their rewards from the verifier's tests. A pass on a wrong run is what \citet{zhang2026leaky} calls a verifier false positive in RLVR.

\paragraph{Postconditions and frame conditions.} In these benchmarks, a task's postcondition says that each requested record exists with the requested values, and its frame condition says that nothing else changes. The frame condition therefore also rules out a second copy of a requested record. Specifications often leave the frame condition out, because listing everything that must stay the same is hard (the frame problem \citep{mccarthy1969philosophical,borgida1995frame}), and a verifier inherits the problem, since checking the frame condition in full means comparing the whole end state with the expected one.

\paragraph{Mutation testing.} For readers who know mutation testing \citep{jia2011mutation}, our edit is a mutation operator, an edited run is a mutant, and a \emph{harmless edit}, one that leaves the result correct such as a repeated call that changes nothing, is an equivalent mutant. Telling a harmless edit from a gap in the tests is hard in general. Here the verifier is the test suite and an agent's run is the program, and the end-state check tells the two apart, so it acts as the oracle \citep{barr2015oracle}. A pass on a harmless edit is correct.

\paragraph{Adversary model.} The adversary knows the task, controls the run, acts only through the benchmark's normal actions from a clean start, and cannot modify the verifier, the evidence the verifier reads, or the environment after the run starts. The model excludes reward tampering, in which an agent changes the reward process itself \citep{everitt2021tampering}. No audit run in this paper used a modified verifier, because a finding has to describe the verifier that benchmark users and RL pipelines run. The repaired copies in \S\ref{sec:repairs} and \S\ref{sec:perstep} are reported apart.

\section{Method}
\label{sec:method}

Our audit takes a benchmark and returns the runs that its verifiers pass although the result is wrong, each confirmed by an end-state check (Figure~\ref{fig:method}).\label{sec:human} After a person writes the edits and a harness for each benchmark, five steps run with no per-run human decision and no model call: listing the edited runs, replaying them, the end-state check, the verifier calls, and classification. Not every step ran in every experiment, and Appendix~\ref{app:roles} says which ran where.

\subsection{Two kinds of edit and the wrong-task test}
\label{sec:operators}

An edited run applies one edit to a correct run (Table~\ref{tab:edits}). Edits of the first kind test whether the verifier checks the requested work, and edits of the second test whether it notices an unrequested change. The wrong-task test grades one task's correct run with another task's verifier, to test whether the verifier checks which task was done. We record a pass on it as an acceptance by the wrong task's verifier. It does not count as a false positive, because no end-state check shows that the other task stayed undone. We designed each edit after reading the verifier and driver source. The edits are not a sample of agent errors or of verifier defects, so a rejection shows only that the verifier checks that edit.

\begin{table}[t]
\centering
\caption{The two kinds of edit and the wrong-task test, which grades a correct run with another task's verifier. Appendix~\ref{app:grammar} defines each edit.}
\label{tab:edits}
\footnotesize
\begin{tabular}{@{}p{2.0cm}p{4.0cm}p{3.2cm}p{3.4cm}@{}}
\toprule
\raggedright Kind of test & \raggedright AppWorld (API calls) & \raggedright WorkArena (browser) & \raggedright A pass means \tabularnewline
\midrule
\raggedright Missing or wrong work & \raggedright claim success and write nothing; drop the last write; keep only the first of a repeated write; change one written value & \raggedright claim success and write nothing; leave the form unsubmitted; leave out a task field & \raggedright a false positive, unless the record is still correct \tabularnewline
\addlinespace
\raggedright Unrequested change & \raggedright repeat the last write & \raggedright also fill a field the task never mentions & \raggedright a false positive, unless the change alters nothing \tabularnewline
\addlinespace
\raggedright Wrong task & \raggedright grade one task's correct run with another task's verifier & \raggedright the same & \raggedright an acceptance by the wrong task's verifier \tabularnewline
\bottomrule
\end{tabular}
\end{table}

\subsection{Replay and end-state check}
\label{sec:roles}

The harness replays each run from a clean start: a reset database in AppWorld, and a fresh ServiceNow episode, cleaned up afterwards, in WorkArena. Before the verifier is called, the end-state check reads what the run changed. It may read the same database as the verifier, but its queries and comparisons are separate from the verifier's code. In AppWorld, the check counts the writes to each table and compares them with the correct run's. That was enough here, because every AppWorld pass on an edited run came from a repeated write. In WorkArena, the paired experiment (\S\ref{sec:exp-workarena}) reads the saved record and compares every task field and the extra field. The benchmark's verifier then grades the run, unmodified, at a fixed version (Appendix~\ref{app:repro}).

\subsection{Outcomes}
\label{sec:confirm}

A fixed rule gives each graded run one outcome. The run is rejected when the verifier fails it. It is a harmless edit when the verifier passes it and the end state is correct. A pass on an edited run is a false positive only when the end-state check, taken before the verifier call, shows the wrong change that the edit names. Any other pass is not counted as a false positive, and no model output enters the rule. Runs that an edit could not apply to, runs that could not be replayed, and runs whose correct run failed are reported apart (Appendix~\ref{app:procedure-full}). Each experiment's edits, task manifest and classification rule were fixed before its graded runs, except one AppWorld write-count comparison added after the first repeated-write verdicts; Appendices~\ref{app:registration} and~\ref{app:evidence-correction} give the chronology, the unresolved runs and a runner fix from an earlier draft. To test a repair, we patch a copy of the verifier by hand and grade the same runs again, or rerun them when the check runs inside the page. The repair works when the edited run now fails and its correct run still passes.

\subsection{When the method applies}
\label{sec:preconditions}

We claim the method only for benchmarks that provide six things (Appendix~\ref{app:preconditions}): start states that can be reset and that a correct run and its edited run can share; a correct run that the verifier passes; programmatic access to the agent's actions; a deterministic verifier that can be called unmodified and graded again as a patched copy; an end-state check that sees each edit's change; and a way to pin what the verifier expects. Each new benchmark needs its own harness, and we have not run it on a third benchmark.

\section{Experiments}
\label{sec:experiments}
\label{sec:audit}

\subsection{Setup}
\label{sec:setup}

AppWorld \citep{trivedi2024appworld,appworld013software} simulates everyday apps, such as a payment app, a music player, notes and a file system, which agents operate through API calls, and grades each task with unit tests over the changes in the app database. We audited 48 AppWorld tasks: 16 task templates whose solutions change data, in three versions each. Eight of the 16 are in AppWorld's official train split and eight in its official dev split; none is in a test split.

WorkArena \citep{drouin2024workarena,workarena053software} is a benchmark of browser tasks on ServiceNow, an enterprise IT platform, whose verifiers check the record that a submitted form saves or the request that a service-catalog order creates. We used the 13 single-step tasks that change data and that the hosted instance's fair-use terms permit, 4 forms and 9 catalog items, in three configurations each. Except under the per-step validation of \S\ref{sec:perstep}, each run calls the verifier once, at the end of the run. BrowserGym, the environment that runs WorkArena agents, calls it after every step in its agent loop \citep{drouin2024workarena}. Appendix~\ref{app:repro} gives versions and the audit set.

\subsection{AppWorld's tests pass a repeated write in two task templates}
\label{sec:exp-appworld}

The edit that adds an unrequested change repeats the last write of the correct run, and AppWorld's tests passed 39 of its 48 runs. In 33 of these, the end-state check found no extra write, because the repeated call changed nothing, so those edits were harmless and their passes correct. That leaves 6 false positives, in all three versions of two task templates. In the work-dinner payment task, the repeated last request leaves an extra request and an extra notification. In the song-queue task, the same song is queued a second time. Neither template's tests count what the run added. The payment tests match new requests to expected shares by receiver (Figure~\ref{fig:example}), and the song-queue tests compare the queue's songs as a set, which cannot see a song queued twice.

A replay that never called the verifier repeated one write in each of the 48 tasks and counted the writes to each table, to find where a repeat can change data (Appendix~\ref{app:eligibility}). In 11 of the 16 templates the repeat changed nothing, so a repeat could only be a harmless edit there. In the other 5, the repeat changed the stored data a second time. The tests of 2 passed it, and the tests of 3 rejected it. The replay detects a repeat only when it changes a write count.

\subsection{A count assertion rejects the six duplicates and passes the correct runs}
\label{sec:repairs}

For the work-dinner payment task, we added one assertion to a copy of the tests: the number of new payment requests must equal the number of coworkers who still owe a share. For the song-queue task, the added assertion requires the queue to grow by exactly the number of songs added. Grading the same runs again failed all 6 edited runs and passed all 6 correct runs, in 12 regrade runs. The example tests in AppWorld's guide for task authors use count checks of this kind on the changed records \citep{appworldgeneratorsguide}. We then tested the copies for false rejections on agent runs. On the same two templates, the original tests accept 13 of 17 archived in-house trials from two models on 4 task ids, and the count assertion rejects none of those 13 natural agent runs. Twelve of the 13 make exactly the correct run's change.

\subsection{WorkArena's form verifier, called after submit, passes an extra field}
\label{sec:exp-workarena}

In WorkArena, the edit that adds an unrequested change fills one field that the task never mentions, the extra field, with a value that differs from the field's form default. We paired each such edited run with a correct run that fills only the task fields. A scan that never called the verifier listed the candidate fields, and 28 remained after a declared rule dropped some (Appendix~\ref{app:evidence-correction}). Each candidate got three pairs, one per seed, 84 pairs in all.

Thirty-six pairs were complete and valid. In each one, the edited run saved the requested value in the extra field, every task field matched, and the verifier passed both runs. All 57 valid correct runs passed. The 36 pairs cover 12 candidate fields in two form templates, incident and change request. The other 48 pairs are unresolved and count as neither acceptance nor rejection. In 27 hardware-asset pairs, our check that the verifier's configuration stays fixed failed, because WorkArena writes a fresh serial number into it (Appendix~\ref{app:evidence-correction}). In the other 21, the edited run stopped at a hidden, read-only, or disabled input before the verifier ran.

The form verifier's documentation says it checks only whether the task's fields have the right values \citep{workarena053software}. Its one check on other fields is a \emph{monitor}, a page script whose source comment reads ``Ensure that only the expected fields are changed''. The monitor sets a flag in the page when a field outside the task changes, and the verifier returns reward 0 when the flag is set (Appendix~\ref{app:monitor}). In all 147 episodes of the paired experiment that reached the verifier, the flag was unset when the verifier read it, and no run of the paired experiment read it earlier.

\subsection{Per-step validation rejects some extra fields and pays others}
\label{sec:perstep}

The monitor keeps its flag in the page, so a verifier call before submit can see it and the page load after submit can lose it. A separate experiment reran the 12 extra fields with one seed on another ServiceNow instance of the same release (Appendix~\ref{app:monitor}). It gave each field three arms, each with its own correct run. A probe fills the extra field last and reads the flag just before the click on Insert that submits the form. Per-step validation calls the verifier after every step, as BrowserGym does, in one run with the extra field filled first and in another with it filled last. The third arm tests a repair, a copy of the verifier that keeps the flag.

\begin{table}[t]
\centering
\small
\caption{Verifier verdicts by kind of test (Table~\ref{tab:edits}), in runs. A pass is harmless when the end state was correct, and a false positive when the end-state check showed the wrong change. Every false positive came from an unrequested change.}
\label{tab:operator-class}
\setlength{\tabcolsep}{3pt}
\begin{tabular}{lrrrrrrrr}
\toprule
 & \multicolumn{4}{c}{AppWorld} & \multicolumn{4}{c}{WorkArena} \\
\cmidrule(lr){2-5}\cmidrule(lr){6-9}
 & & & \multicolumn{2}{c}{Passed} & & & \multicolumn{2}{c}{Passed} \\
\cmidrule(lr){4-5}\cmidrule(lr){8-9}
Kind of test & Graded & Rejected & harmless & false pos. & Graded & Rejected & harmless & false pos. \\
\midrule
Missing or wrong work & 168 & 168 & 0 & 0 & 93 & 84 & 9 & 0 \\
Unrequested change & 48 & 9 & 33 & 6 & 36$^{\dagger}$ & 0 & 0 & 36 \\
Wrong task & 2{,}263 & 2{,}263 & n/a & n/a & 426 & 426 & n/a & n/a \\
\bottomrule
\end{tabular}
\par\vspace{2pt}\begin{minipage}{\linewidth}\footnotesize\raggedright
$^{\dagger}$The edited runs of the 36 complete and valid pairs of the paired experiment, which calls the verifier once after submit (\S\ref{sec:exp-workarena}); each one's correct run passed. The other 48 of the 84 scheduled pairs were unresolved. Wrong-task runs exclude each task graded by its own verifier, which passed every time (31 of 31 and 39 of 39), and 3 WorkArena runs that errored. The per-step experiment (\S\ref{sec:perstep}) is not in this table.
\end{minipage}
\end{table}

\begin{table}[t]
\centering
\footnotesize
\caption{Per-step validation in the separate per-step experiment (one seed), by the kind of extra field and when it was filled. A drop-down is a select field, and a reference field links to another record. One correct run was unresolved, so the counts cover 11 of the 12 fields. Table~\ref{tab:wa-perstep} gives every arm.}
\label{tab:perstep}
\begin{tabular}{@{}llrl@{}}
\toprule
Extra field & Filled & Runs & Outcome under per-step validation \\
\midrule
any kind & first & 11 & reward 0 before submit \\
drop-down & last & 4 & reward 0 before submit \\
text or reference & last & 7 & reward 1 at submit, as a correct run \\
\midrule
none (correct run) & & 11 & reward 1 at submit \\
\bottomrule
\end{tabular}
\end{table}

The experiment ran 84 episodes, and 83 were valid. The end-state check confirmed the extra value in every valid edited run. Under per-step validation, the outcome depended on the kind of field and on when it was filled (Table~\ref{tab:perstep}).

The probe shows why: all 12 edited runs passed after submit, as in the paired experiment. A drop-down set the flag before submit in all 4, and the page load after submit lost it. In the 8 probe runs with a text or reference field, the flag was unset just before submit, as in all 12 correct runs. Filled last, such a field set the flag only during the submit click, so the same page load lost it. Filled first, it set the flag when the runner filled the next field. The flag-keeping copy rejected all 12 edited runs and passed all 12 correct runs (Appendix~\ref{app:monitor}). A second repair would compare the created record's other fields with their defaults; we propose it but did not run it.

\subsection{The same verifiers catch the missing and wrong work we planted}
\label{sec:cross-benchmark}

The same verifiers rejected 252 of the 261 graded runs with missing or wrong work (Table~\ref{tab:operator-class}): 168 of 168 in AppWorld and 84 of 93 in WorkArena. The 9 they passed are harmless edits. Each left out a task field whose form default is the requested value, so the record is correct. They passed none of the 2{,}689 valid wrong-task runs, 118 of them between tasks that differ in one parameter (Appendices~\ref{app:grids} and~\ref{app:session}).

\section{What a test-based reward pays}
\label{sec:reward}

The work-dinner payment task's three versions are 3 of the 90 tasks in AppWorld's official train split, and the song-queue task is in the dev split; neither is in a test split (Table~\ref{tab:reward}). CANOPY pays 1 only when every test passes \citep{pu2026canopy}, and Entropy-Preserving RL pays the fraction of tests that pass \citep{petrenko2026entropy}. ProCredit adds to the all-tests reward a progress term: a constant $c > 0$ times the gain, over the run, in the fraction of initially failing tests that pass \citep{ma2026procredit}. All three train on all 90 training tasks. LOOP \citep{chen2025loop}, GVPO \citep{dai2026gvpo}, and SAGE \citep{wang2025sage} train only on difficulty-1 and difficulty-2 tasks, which leaves out the difficulty-3 payment task, and select checkpoints on the dev split, which holds the song-queue task (for LOOP, which does not list its dev tasks, this is our inference).

\begin{table}[t]
\centering
\footnotesize
\caption{Where the two AppWorld task templates with false positives sit in RL pipelines, for all three versions of each. We derived the rewards from the verdicts and each method's formula.}
\label{tab:reward}
\setlength{\tabcolsep}{4pt}
\begin{tabular}{@{}p{2.0cm} p{1.45cm} p{3.2cm} p{2.6cm} p{3.6cm}@{}}
\toprule
\raggedright Task template & \raggedright Official split & \raggedright Trains on it & \raggedright Selects checkpoints on its split & \raggedright Duplicate vs.\ correct run \tabularnewline
\midrule
\raggedright work-dinner payment & \raggedright train, difficulty 3 & \raggedright CANOPY, Entropy-Preserving RL, ProCredit & \raggedright none; selection uses dev & \raggedright both pass 4 of 4 tests; the duplicate adds one payment request and one notification \tabularnewline
\addlinespace
\raggedright song queue & \raggedright dev, difficulty 1 & \raggedright none & \raggedright LOOP, GVPO, SAGE, Entropy-Preserving RL & \raggedright both pass 6 of 6 tests; the duplicate adds one song-queue write \tabularnewline
\bottomrule
\end{tabular}
\end{table}

In all six duplicate-write runs, the duplicate passed the same tests as its correct run, although the end-state check shows its extra write. By our derivation from the published formulas, each duplicate therefore earns its correct run's reward under all three, for any $c > 0$. Nothing in these rewards penalizes the extra write.

AppWorld's released 2024 baseline outputs for its two test splits hold 1{,}277 passing runs across 28 agent and model configurations, and none of them repeats the payment-request create call or the song-queue add call with the same payload. The loophole is therefore latent in these released runs, and whether RL training would bring the repeats out, for example after a lost acknowledgement \citep{li2026exactlyonce}, is open. We audited 8 of AppWorld's 30 official-train task templates and none of the other 22.

On WorkArena, \citet{vattikonda2025webagent} trained policies with group relative policy optimization (GRPO) on 24 tasks, among them the incident, change-request, hardware-asset, and problem creation forms we audited. They use BrowserGym but do not say whether their reward comes from its per-step validation. In a separate experiment on 12 incident and change-request fields with one seed, per-step validation gave a correct run's reward only to edited runs that filled a text or reference extra field last (\S\ref{sec:perstep}).

\ifpilot
\input{sections/checkers/12_pilot}
\fi
\section{Related work}
\label{sec:related}

\paragraph{Verifier error in RL.} RLVR still learns from a noisy verifier as long as its true-positive rate exceeds its false-positive rate \citep{rad2026rlver}. Systematic false positives, by contrast, can stall or collapse training, depending on the pattern of errors more than their overall rate \citep{egashira2026delayplateaucollapseevaluating}. A loophole that holds across a task template, like ours, is systematic. In Agent World Model, a suite of synthetic RL environments for agents, 16.8\% of 10{,}000 machine-written verifiers pass an agent that does nothing, and making half the training verifiers over-permissive costs 7 points of true capability \citep{anon2026doseresponse}.

\paragraph{Automated red-teaming and reward-hacking audits.} BenchJack, whose exploits reach near-perfect scores on 9 of 10 agent benchmarks without solving a task \citep{wang2026benchjack}, and a hacker-fixer loop, which finds 323 of 1{,}968 terminal-agent tasks hackable by frontier models \citep{zhong2026hackerfixer}, aim learned attacks at benchmark verifiers. RewardProof validates 1{,}116 attacks in 567 environments by comparing the reward with and without each attack \citep{anon2026rewardproof}, and \citet{luo2026unearned} screen 3{,}810 passing runs with LLM judges. AgentHorizon builds negatives by pairing a recorded run with a closely related instruction, and its best judge, GPT-5.5 in Codex, identifies only 24.1\% of the runs labeled as bad side effects \citep{lu2026agenthorizon}. RECEIPT separates proposing from confirming, with a deterministic replay confirming the XSS exploits an agent proposes \citep{lyu2026receipt}, and PartHackBench keeps an adversarial tool-agent run only when a private certifier finds its progress equal to an honest run's, then measures the extra credit \citep{yang2026parthackbench}. Our audit confirms each finding against the end state, which shows that a run with an unrequested change is wrong although its score and reward do not (\S\ref{sec:intro}).

\paragraph{Audits of benchmark verifiers.} \citet{zhu2025abc} call the property that a benchmark's verdict reflects real task success \emph{outcome validity} and find that $\tau$-bench counts empty responses as successes; a false positive is a failure of outcome validity. \citet{dong2026miscore} find 15.3\% of 150 audited fail verdicts wrong across five computer-use benchmarks, WorkArena among them, and do not audit pass verdicts. \citet{gao2026evidence} attach outcome evidence to 300 natural AppWorld runs and, after fixing their own scorer, find all 220 released successes supported. In concurrent work, grading after every API call turns 17 of 142 published AppWorld successes into failures \citep{anon2026grader}. In code, SWE-ABS, UTBoost, PatchDiff, TestJack, agent-built test suites, and ImpossibleBench catch passing solutions that violate the specification \citep{yu2026sweabs,yu2025utboost,wang2026patchdiff,anon2026testjack,xie2026testsuite,zhong2025impossiblebench}. Concurrent work also tests released verifiers on crafted wrong inputs: the released MathConstruct verifiers accept all 127 applicable crafted invalid constructions \citep{anon2026acceptedinvalid}, and KernelBench's official verifier misses 16.9\% of witnessed faults \citep{du2026measuring}. Gaia2 tests its verifier on perturbed oracle actions, which catch only anticipated behaviors \citep{froger2025are,froger2026gaia2}. Our hand-written edits share that limit.

\paragraph{Unrequested changes.} \citet{sun2026mutation} mutates known-valid final states of $\tau^2$-Bench, ToolSandbox and AppWorld tasks, one state mutation at a time, and re-runs the official evaluator with no model call (\S\ref{sec:intro}). The preprint was posted on 18 August 2026, before this paper was submitted. Sun's AppWorld sample had no applicable collateral mutation. Our edits, executed through AppWorld's API, reached the duplicate request that the snapshot method could not (\S\ref{sec:exp-appworld}). Across 25{,}930 sandbox episodes, \citet{li2026exactlyonce} measures how often agents duplicate a write after a lost acknowledgement, a natural source of the repeats we plant. Agent-Diff grades a new benchmark of enterprise API tasks by the state difference under a closed-world invariant, so a change the task did not call for fails the run \citep{pysklo2026agentdiff}, and EffectMatch compares an agent's persistent changes at run time with the approved outcome and blocks the commit on a mismatch \citep{zhang2026effectmatch}. Neither tests an existing verifier. WorkArena issue \#155 reports that the delete-record verifier accepts deleting unrelated records \citep{kimjune2026issue155}, and \citet{bellibatlu2026contract} check 34 tools in four tool-use benchmarks against contracts with frame clauses. We test shipped verifiers on executed runs and confirm each pass against the end state.

\section{Limitations}
\label{sec:limitations}

\emph{What the counts can show.}\label{sec:certificate}
A false positive shows that a verifier is wrong on one run. Our audit cannot show that a verifier is sound, cannot find a loophole that no edit reaches, and tests false rejections only through the correct runs and the patched copies. Verdicts within a task template almost never differ (in Table~\ref{tab:fault-flow}, 107 of the 109 template-by-edit groups with more than one applicable run return one verdict throughout), so a binomial interval over runs would count repeated versions as independent observations. The counts give no population rate, which would need a sampling frame fixed before outcomes are visible and an independent holdout. A sealed holdout of 14 AppWorld task templates, which no run in this paper opened, stays sealed for that study.

\emph{Two benchmarks, hand-written edits.} We used no model-generated adversary, adaptive search, or corpus of natural failures, and in AppWorld the end-state check counted writes and compared no values (\S\ref{sec:roles}), so the counts are a lower bound (\S\ref{sec:intro}).

\emph{WorkArena coverage.} In the paired experiment, 36 of 84 pairs were complete and valid, in two of the four form templates. Its sampling frame did not depend on earlier verdicts. After delivery limits, though, its 12 candidate fields are the same ones an earlier rerun covered, whose verdicts we had seen (Appendix~\ref{app:evidence-correction}). The paired experiment's results hold for one verifier call after submit. The per-step experiment used one seed per field and another instance, placed the extra field only first or last, and at submit used the paired experiment's verifier call in place of BrowserGym's.

\emph{Reproducibility.} Appendix~\ref{app:repro} gives pins and reset boundaries, and a checksum-bound supplement accompanies the paper (Appendix~\ref{app:artifacts}). No run in this paper calls a model. The 13 natural runs of \S\ref{sec:repairs} come from earlier in-house work. \TODOdisclosure

\makeatletter
\@ifundefined{ifpilot}{\expandafter\newif\csname ifpilot\endcsname}{}
\makeatother
\section{Conclusion}
\label{sec:conclusion}

A benchmark verifier is a test suite for agent behavior, so mutation testing can check it. In both AppWorld and WorkArena, our audit found a side-effect loophole: a run that did the task and made one unrequested change still passed. Our edits that left work missing or wrong produced no false positive, so the gap we found lies in the frame condition. Both benchmarks do include a check meant for unrequested changes, in AppWorld's tests and in a page script on WorkArena's forms, but these runs got past both. Because AppWorld's tests pass the run that repeats a payment request, the reward formulas of three recent RL methods give it the correct run's reward. One count assertion in each of the two affected AppWorld templates rejected the duplicate in all six tasks and failed no correct run.

Two questions remain open. The first is whether RL optimization finds the duplicate on its own, which training policies on the original and on the repaired AppWorld tests would answer. The second is how far automated auditing can replace hand-written edits: when a language model reads a verifier and proposes candidate loopholes, the bottleneck shifts to confirming which candidates produce an unauthorized end state and repairing the verifier without rejecting valid alternative runs (Appendix~\ref{app:proposer}).\ifpilot{} \pilotfuturepointer{}\fi{} Before a verifier becomes a reward, its authors can count what a run adds or compare the whole end state with the expected one.

\begin{ack}
The author is the founder and CEO of MLDSAI Inc. (OpenAdapt.AI), which
develops open-source GUI automation software, including tools that check
whether an automated action had its intended effect. This work received no
external funding.
\end{ack}

\bibliography{references_checkers}
\bibliographystyle{abbrvnat}

\appendix

\section{Edit definitions}
\label{app:grammar}

\paragraph{Notation.} A benchmark task $t$ has a start state $s_0$ and an action channel: API calls in AppWorld, and GUI actions through the shipped task driver in WorkArena. The task's verifier $V_t$ returns \textsc{Pass} or \textsc{Fail}, and in some configurations it abstains from scoring. The effect of a run is the typed difference between its start and end states, as records added, removed, and updated. Effects are compared after dropping a declared set $N$ of fields with no task meaning, such as ids, timestamps, auto-numbers, and display aliases. The method's comparator treats two effects as equal when their added, removed, and updated records agree as multisets after dropping $N$. The AppWorld edit runs compared write counters instead, which count the writes to each table but not their values. Template \texttt{22cc237} is the work-dinner payment task, and \texttt{396c5a2} is the song-queue task.

\paragraph{Edits.} Each kind of edit covers one way an effect can be wrong: it lacks part of the request, adds something the request did not ask for, or is the correct effect of a different task. We designed each edit after reading the verifier and driver source. Two combinations have no edit: no API edit writes a field or record the task never names, and no GUI edit repeats a submit. Table~\ref{tab:grammar-defs} defines each edit. Three edits pick their targets from a scan: \textsc{Extra-NI} from the idempotence replay, \textsc{Omit-Divergent} from the on-form defaults, and \textsc{Extra-Field} from candidates whose template value differs from the form default. Each scan's targeting was frozen in a commit before the runs it enables were scored. A run enters the false-positive count only when the end-state check confirms the named wrong effect.

\begin{table}[h]
\centering
\caption{Edits. Each edited run applies one edit and is delivered from a fresh start state.}
\label{tab:grammar-defs}
\small
\begin{tabular}{@{}l l p{6.7cm}@{}}
\toprule
Edit & Kind & What it does \\
\midrule
\textsc{Claim-Only} & missing or wrong work & Deliver only the last read or claim call, so no write occurs. \\
\textsc{Omit} & missing or wrong work & Drop the last write call of the correct run's program. \\
\textsc{Partial} & missing or wrong work & Keep only the first of a repeated write sequence. \\
\textsc{Wrong-Value} & missing or wrong work & Move one written value to a different legal value. \\
\textsc{Extra} & unrequested change & Repeat the last write call. \\
\textsc{Extra-NI} & unrequested change & \textsc{Extra}, limited to writes whose repeat changes the write counters in a committed idempotence replay. Added after the first \textsc{Extra} run; it re-scores the same 15 edited runs. \\
\textsc{Claim-Only} (GUI) & missing or wrong work & Claim completion without creating the underlying request. \\
\textsc{No-Submit} (GUI) & missing or wrong work & Leave the form unsubmitted, and do not invoke the task driver. \\
\textsc{Omit-Field} (GUI) & missing or wrong work & Submit the form without one task field. \\
\textsc{Omit-Divergent} (GUI) & missing or wrong work & Omit a task field whose requested value differs from its on-form default, as found by a committed default scan. \\
\textsc{Extra-Field} (GUI) & unrequested change & Also write a field the task never names, with a value that differs from the field's form default (for example \texttt{category} or \texttt{contact\_type} on the incident form). \\
wrong-task test & wrong task & Deliver task $j$'s correct run to task $i$'s verifier, labeled by distance level: one-parameter twin, same task template, another task template with shared tables, or one with disjoint tables. \\
\bottomrule
\end{tabular}
\end{table}

\section{Protocol freezes and interruptions}
\label{app:registration}

Each experiment had three stages: the protocol and enumeration code, the required fixed run sets, and the scored runs. The commit that adds a stage precedes the artifacts it governs. These commits are private; they establish chronology and are not a public preregistration. External process kills and one session loss interrupted several runs. Each resumption skipped every recorded run, so no graded run was scored twice, and the ledgers record each interruption. Two diagonal shakedown runs preceded the earliest WorkArena freeze; the protocol discloses them, and the scored grid re-ran both. Refusals are recorded with their reasons. The paired experiment's record check was committed before its first scored run. The AppWorld counter comparison was committed after the first \textsc{Extra} verdicts and before the \textsc{Extra-NI} re-scores (\S\ref{sec:exp-appworld}). No model output enters either. The protocol for testing the AppWorld repair copies on natural agent runs (\S\ref{sec:repairs}) was written before any agent run met them. We added the idempotence replay and \textsc{Extra-NI} after the first \textsc{Extra} run; \textsc{Extra-NI} re-scored the same 15 edited runs, reproduced all 15 verdicts, and adds no run.

\section{Wrong-task grids and the WorkArena selection rule}
\label{app:grids}

The wrong-task test delivers task $j$'s correct run to task $i$'s verifier. When $j = i$ the run is the task's own \emph{diagonal run}; every other pair is \emph{off-diagonal}. Each swap has one of four distance levels, from nearest to farthest: a twin that differs in one parameter, a sibling from the same task template, a task that writes the same tables, or a task that writes disjoint tables. The wrong-task test delivered 2{,}263 valid off-diagonal runs to AppWorld verifiers, 43 of them one-parameter twins. They passed none, and all 31 diagonal runs passed. In WorkArena, the wrong-task test delivered 426 valid off-diagonal runs, 75 of them one-parameter twins. The verifiers passed none, and all 39 diagonal runs passed. We exclude three runs that raised a verifier error. The cross-template runs cover 273 of the 1{,}404 cross-template pairs of task instances, by a rule frozen before scoring (Table~\ref{tab:wa-selection}).

Table~\ref{tab:swap} gives the wrong-task grids by distance level. The AppWorld grid crosses 74 verifiers with 31 freshly executed correct runs. Its frozen protocol fixes the verifiers by rule: 31 are the verifiers of the audit-set tasks in the 19 August 2026 twin-gate run, and 43 are those of the admissible one-parameter twins in the committed gate manifest. The \texttt{22cc237} twins are left out because their twin rule was withdrawn. The grid has 2{,}294 runs, all sequential, with no verifier error or timeout. Grading took about 80 minutes in total, at about 2\,s per run. An external process kill interrupted the grid, and grading resumed from the recorded runs, so the 531\,s in the ledger covers only the final resume pass. The WorkArena grid covers 13 templates at configuration indices 0--2: 468 runs over seven resume passes and about 11 hours, with three verifier errors excluded from the counts. An earlier 6-task AppWorld pilot and a 12-run WorkArena pilot agree with the full grids, which subsume them.

\begin{table}[t]
\centering
\small
\caption{Wrong-task grid: one task's correct run graded by another task's verifier, by distance level. An off-diagonal \textsc{Pass} would be an acceptance by the wrong task's verifier, never a false positive.}
\label{tab:swap}
\setlength{\tabcolsep}{3pt}
\begin{tabular}{llrr}
\toprule
Suite & Distance level & Valid runs & Verifier \textsc{Pass} \\
\midrule
AppWorld (2{,}294 runs) & Diag (own verifier) & 31/31 pass & n/a \\
 & 0: one-parameter twin & 43 & 0 \\
 & 1: sibling, same template & 144 & 0 \\
 & 2: other template, shared tables & 369 & 0 \\
 & 3: other template, disjoint tables & 1{,}707 & 0 \\
\midrule WorkArena (468 runs; 3 errors) & Diag (own verifier) & 39/39 pass & n/a \\
 & 0: one-parameter twin & 75 & 0 \\
 & 1: same template, other configuration & 78 & 0 \\
 & 2: other template, shared tables & 216 & 0 \\
 & 3: other template, disjoint tables & 57 & 0 \\
\bottomrule
\end{tabular}
\par\vspace{2pt}\begin{minipage}{\linewidth}\footnotesize\raggedright
Generated from the committed grid artifacts \texttt{analysis/intent\_swap/fullgrid\_results.json} and \texttt{analysis/intent\_swap/workarena\_grid\_results.json}.
\end{minipage}
\end{table}

The WorkArena grid does not cross every pair of its 39 task instances. Table~\ref{tab:wa-selection} gives the selection rule, which the grid protocol (\texttt{PROTOCOL\_INTENT\_SWAP\_WA\_GRID\_2026\_08\_23.md}) froze after the 12-run pilot on 4 templates, whose 8 off-diagonal runs all failed, and before the first grid run was scored. The rule keeps the two levels nearest the target complete: every twin and every same-template pair. Cross-template runs pair configurations only at equal index, plus a fixed rotation of 21 form-and-catalog runs at index 0, which gives 273 of the 1{,}404 cross-template pairs. We chose the rule to limit run time. The pool permits one episode at a time, and the pilot report, written before the freeze, measured about 71\,s per run, estimated about 60 hours of sequential pool time for the full grid of 3{,}042 runs, and recommended a scoped grid.

\begin{table}[h]
\centering
\caption{WorkArena wrong-task grid: the frozen selection rule (468 runs over 39 task instances, 13 templates at configuration indices 0--2). Cross-template rows total 273 of the 1{,}404 possible cross-template pairs.}
\label{tab:wa-selection}
\small
\begin{tabular}{@{}l l r p{5.5cm}@{}}
\toprule
Run class & Distance level & Runs & Rule \\
\midrule
correct run & diagonal & 39 & each instance with its own verifier \\
moved trajectory & one-parameter twin & 39 & base verifier, trajectory with one parameter moved \\
moved verifier & one-parameter twin & 39 & moved verifier, base trajectory \\
sibling configuration & same template & 78 & every ordered configuration pair within a template \\
catalog pairs & cross-template, shared tables & 216 & all 72 ordered catalog-template pairs at equal index \\
form pairs & cross-template, disjoint tables & 36 & all 12 ordered form-template pairs at equal index \\
cross-family sample & cross-template, disjoint tables & 21 & each form at index 0 with 3 catalogs (rotation $(3i{+}j) \bmod 9$); each catalog at index 0 with form $j \bmod 4$ \\
\bottomrule
\end{tabular}
\end{table}

\section{WorkArena's session-scoped evidence}
\label{app:session}

In \texttt{browsergym-workarena} 0.5.3, a submit-time hook of the form-task verifier stores the created record's system id under a \texttt{localStorage} key that contains the Python object identity of the verifying task instance. During validation the verifier reads that key and returns ``The form has not been submitted'' when it is absent. The catalog verifier instead reads the request id from the current page URL. Probe runs confirmed the mechanism: the target record existed in the database, and a verifier outside the submitting session returned \textsc{Fail}. The benchmark supports grading only within the submitting session, so we count this as a limit on moving runs across sessions and not as a false reject. Fifty-seven wrong-task rejections rest only on evidence kept in the submitting browser session, so they show nothing about task identity.

\section{WorkArena's out-of-scope-field monitor}
\label{app:monitor}

In \texttt{browsergym-workarena} 0.5.3, the form-task setup injects a script that the authors introduce with the comment ``Ensure that only the expected fields are changed''. The script finds every \texttt{input}, \texttt{select}, and \texttt{textarea} element on the page at injection time. For each one whose trailing field name is not in the task's protected list, it registers a \texttt{change} listener that sets a page-local flag, \texttt{WORKARENA\_BAD\_FIELD\_CHANGED}. The verifier reads that flag first and, when it is set, returns reward 0 with ``Some fields outside of the task scope have been changed.'' The docstring of \texttt{validate()} says that its record check covers only whether the expected fields have the right value, not whether extra fields changed, so the flag is its only check on fields outside the task.

The paired-experiment runs fill the task fields and the extra field through the shipped fill routine, submit the form by clicking Insert, read the record back through the Table API, and then call \texttt{validate()} once. In all 147 paired-experiment episodes that reached the verifier (168 scheduled, less the 21 edited runs that stopped before it), the flag was unset when \texttt{validate()} read it. No paired-experiment run read the flag before submit, so the paired experiment cannot tell whether the listener fired and the page load after submit lost the flag, or whether the listener never fired. Listeners bound only at injection time could miss fields created later. The paired experiment admitted only fields present on the form (for example \texttt{category} and \texttt{contact\_type}), and the end-state check confirmed the requested value in the extra field in all 36 valid edited runs of the paired experiment.

\paragraph{Per-step experiment.} The per-step experiment of \S\ref{sec:perstep} addresses that question for the same 12 candidate fields, in new episodes on another instance. We froze its protocol in a commit on 8 October 2026, before its first request to an instance. It schedules seven episodes per field, 84 in all, with seed 20260829, the first of the three seeds it registers: a correct run and an edited run for the flag probe, a correct run and two edited runs for per-step validation (extra field first, extra field last), and a correct run and an edited run for the flag-keeping copy. Every episode runs the paired-experiment episode code inside a harness that wraps four of its helpers, and keeps the paired experiment's driver isolation, configuration hashes, Table API end-state check before the verifier call, and cleanup. The probe reads the flag just before submit. Under per-step validation, the runner fills one field per step, in a fixed order, and after each one calls BrowserGym's own post-step routine with validation on, which is the code that \texttt{env.step} runs after an action. Submit is the last step; at submit, the paired experiment's single verifier call, made after the Table API end-state check, stands in for BrowserGym's post-step call. Before submit no record exists, so \texttt{validate()} returns reward 0 at every step, and it ends the episode only when the flag is set. An episode's reward is 0 if a step before submit ended it, and otherwise the reward after submit. The runner keeps delivering after a step ends the episode, so every run under per-step validation also has a verdict after submit. The flag-keeping copy is a subclass of the shipped task class whose monitor also writes the flag to the episode browser's \texttt{localStorage} and whose \texttt{validate()} rejects the run when that stored flag is set; the installed package is not edited. It installs its own monitor, so it ran in new episodes. By 8 October the paired-experiment instance had left the pool, so an amendment, which the author approved and we committed before any request under it, pinned the first pool instance that passed the paired experiment's release checks. Before the campaign, three unscored correct-run episodes, one per arm, checked the harness; they are not counted. The campaign ran on 8--9 October 2026 (UTC). One correct run under per-step validation was unresolved, which leaves 35 of the 36 groups complete; an unresolved episode counts as neither an acceptance nor a rejection. Table~\ref{tab:wa-perstep} gives the counts by field kind. For text and reference fields filled first, the episode ended one step after the write, when the runner filled the next field. The two valid edited runs of the incomplete group, on a text field, behaved like the others: the one that filled it first ended one step after the write, and the one that filled it last received reward 1 at submit. After submit, the shipped verifier passed 10 of the 11 edited runs in complete groups that filled the extra field first and 3 of the 4 that filled a select field last, and the end-state check confirmed the extra value in each; in the other two, both on the incident \texttt{category} field, the flag was still set after submit and the verifier rejected the run.

\begin{table}[t]
\centering
\small
\caption{WorkArena per-step experiment, by the kind of the extra field: select (drop-down), text, or reference (a link to another record). Each field ran three arms, each with its own correct run: the flag probe, per-step validation, and the flag-keeping copy. Of 84 scheduled episodes, 83 were valid; the counts cover the 35 of 36 groups (one field in one arm, with its correct run) that were complete and valid. One correct run under per-step validation, for a text field, was unresolved, so the per-step validation columns count 4 of the 5 text fields. Flag set: in the flag probe, the monitor's flag was set when read just before the click on Insert that submits the form. Per-step validation: after each field, BrowserGym's post-step routine calls the verifier, and at submit the paired experiment's single verifier call does. Ended: the verifier ended the episode with reward 0 before submit, with the extra field filled first or last. Reward 1 at submit: no step before submit ended the episode, and the verifier passed the run after submit; edited-run rows count the runs that filled the extra field last. Flag-keeping copy: a copy of the verifier that keeps the flag across the page load after submit. The counts describe these 12 fields and give no rate.}
\label{tab:wa-perstep}
\setlength{\tabcolsep}{3pt}
\begin{tabular}{lrrrrrr}
\toprule
& & Flag probe & \multicolumn{3}{c}{Per-step validation} & Flag-keeping \\
\cmidrule(lr){3-3}\cmidrule(lr){4-6}\cmidrule(lr){7-7}
Field kind & Fields & flag set & ended, first & ended, last & reward 1 at submit & copy rejects \\
\midrule
select & 4 & 4/4 & 4/4 & 4/4 & 0/4 & 4/4 \\
reference & 3 & 0/3 & 3/3 & 0/3 & 3/3 & 3/3 \\
text & 5 & 0/5 & 4/4 & 0/4 & 4/4 & 5/5 \\
\midrule correct runs & 12 & 0/12 & \multicolumn{2}{c}{0/11} & 11/11 & 0/12 \\
\bottomrule
\end{tabular}
\end{table}

\section{WorkArena correction and run coverage}
\label{app:evidence-correction}

\paragraph{Change note.} This version replaces the reviewed WorkArena result, which reported false positives in three form templates from a rerun of 23 runs selected because they had already passed, with a paired experiment: 36 of 36 complete valid pairs over two form templates.

\paragraph{History.} The reviewed version of this paper reported a rerun of the 23 extra-field runs that had already passed. The Table API found a nondefault value in 21 of them, and the verifier passed all 23. We no longer count that rerun, because its driver task object was built from the same configuration object as the verifier's task object, and WorkArena's task constructor writes per-object unique values into that configuration. In the 9 hardware-asset runs, the driver therefore replaced the verifier's expected serial number with its own, so the saved record matched the replaced value and not the serial number in the goal. Under the original expectation those 9 records would have failed on serial number. The extra field itself went into a new copy of the driver's field list, so the verifier's task and protected field lists did not change. The 12 confirmed incident and change-request runs have no unique-valued field and are unaffected. The other 2 runs were harmless edits whose requested display label maps to the stored default. An offline object-identity probe reproduces the overwrite in 9 of the 9 hardware runs, in none of the other 14, and in none of the 23 when the driver gets a deep copy.

\paragraph{Earlier WorkArena runs.} The defect cannot have produced any \textsc{Fail} verdict in the missing-or-wrong-work or wrong-task runs. The missing-or-wrong-work runs did share one configuration object between driver and verifier, but their runner already set the driver's serial number to the verifier's expected value, so the overwrite could only turn a serial-number mismatch into a match. The wrong-task runs build driver and verifier from separate configuration objects, and the diagonal runs use no separate driver. None of the 9 missing-or-wrong-work runs that passed is a hardware-asset run.

\paragraph{The first extra-field experiment.} \textsc{Extra-Field} writes one field the task never names, with a value that differs from the field's form default. The first \textsc{Extra-Field} experiment registered 76 runs: 40 had no eligible extra input, 13 failed before submit, and 23 reached submit and passed. It never checked the extra field, so we ran a paired experiment.

\paragraph{Paired-experiment coverage.} A field scan that never called the verifier listed 36 candidate fields on the 12 form-task configurations, whether or not an earlier run had passed. These 36 extra-field candidates were the paired experiment's sampling frame. A first attempt stopped after 37 episodes, all of which received \textsc{Pass}. A declared rule dropped 8, and we committed the 28-candidate manifest about ten minutes before the first paired-experiment episode. Of the 8, in 3 the requested value mapped to the stored default, and in 5 the default was an auto-number or a timestamp. Because all 37 earlier episodes had passed, no verifier rejection could have shaped those exclusions. The manifest commit is dated 04:19:59 UTC on 29 August 2026. Each candidate got three pairs, with seeds 20260829 to 20260831; the correct run fills only the task fields, and the edited run also fills the candidate as its extra field. The paired experiment scheduled 84 pairs and 168 episodes, which started between 04:30 and 16:31 UTC that day. Ninety-three episodes are valid: 57 correct runs and 36 edited runs. The 54 hardware-asset episodes fail the frozen configuration check, because WorkArena's task constructor writes a fresh serial number into each task object's configuration. Twenty-one edited runs for seven problem and change-request candidates stopped before the verifier ran, because Playwright found the target input hidden, read-only, or disabled. Those seven candidates, two problem and five change-request fields, are the seven of the 28 that were not among the 23 earlier runs that had passed. The 36 complete valid pairs cover 12 candidate fields over seven field names in two templates, incident (24 pairs) and change request (12): \texttt{category}, \texttt{contact\_type}, \texttt{description}, \texttt{close\_notes}, \texttt{cmdb\_ci}, \texttt{assignment\_group}, and \texttt{backout\_plan}. These are the same 12 candidate fields that the configuration overwrite did not affect in the earlier 23-run rerun, whose verdicts we had seen; the same delivery limits filtered both sets. The paired experiment adds paired correct runs, three seeds, driver and verifier isolation, and a full task-field end-state check, but no new field.

\paragraph{Runner integrity.}
\label{sec:readback}
The integrity checks of step~5 (Appendix~\ref{app:procedure-full}) exist because of a defect in our runner. In the reviewed version, the WorkArena driver shared one configuration object with the verifier and overwrote the serial number the verifier expected in 9 hardware-asset runs, all of which we had counted as confirmed. We found the defect later, by reading the code. The paired-experiment runner builds the driver from a deep copy of the configuration, checks that driver and verifier share no mutable object, and hashes what the verifier expects at four stages from reset through validation (Appendix~\ref{app:repro}). That hash check also rejected all 54 hardware-asset paired-experiment episodes, because WorkArena's task constructor writes a fresh serial number into the verifier's configuration. It would reject these 54 with or without our defect, so they cannot show that it would have caught the defect.

\section{Edit run flow and duplicate-write eligibility}
\label{app:fault-flow}

Table~\ref{tab:fault-flow} gives the run flow for every edit, with construction and delivery losses shown before the verdicts. The 33 \textsc{Extra} runs it counts as harmless edits are the 33 runs its \textsc{Extra-NI} note counts as refused: the idempotence replay refused their tasks because no repeated write changed a write counter.

\begin{table}[t]
\centering
\scriptsize
\caption{Run flow for each distinct edit. Scheduled runs (Sched.) partition into refused or ineligible edits (Ref.), incomplete deliveries (Inc.), harmless edits (Harml., edits that left the effect correct), unconfirmed candidates (Cand.), and classified runs (Class.). Class. repeats the frozen report's classification, and not every run in it has independent state evidence. Pass and FP count outcomes and are not part of that partition. FP counts false positives, so a zero means only that none was confirmed. The upper \textsc{Extra-Field} row is the first \textsc{Extra-Field} experiment, which never checked the extra field, so its 23 passes stay candidates and its FP is 0. The lower panel ($\dagger$) is the paired experiment of Table~\ref{tab:faults}, counted in pairs of a correct run and an edited run. Integ.\ counts hardware-asset pairs that failed the frozen configuration check. Stop.\ counts edited runs that stopped before the verifier ran. The last three columns count complete valid pairs.}
\label{tab:fault-flow}
\setlength{\tabcolsep}{3pt}
\begin{tabular}{lrrrrrrrr}
\toprule
Benchmark:edit & Sched. & Ref. & Inc. & Harml. & Cand. & Class. & Pass & FP \\
\midrule
appworld:\textsc{claim-only} & 48 & 0 & 0 & 0 & 0 & 48 & 0 & 0 \\
appworld:\textsc{extra} & 48 & 0 & 0 & 33 & 0 & 15 & 39 & 6 \\
appworld:\textsc{omit} & 48 & 0 & 0 & 0 & 0 & 48 & 0 & 0 \\
appworld:\textsc{partial} & 48 & 9 & 0 & 0 & 0 & 39 & 0 & 0 \\
appworld:\textsc{wrong-value} & 48 & 15 & 0 & 0 & 0 & 33 & 0 & 0 \\
\midrule workarena:\textsc{claim-only} & 52 & 25 & 0 & 0 & 0 & 27 & 0 & 0 \\
workarena:\textsc{extra-field} & 76 & 40 & 13 & 0 & 23 & 0 & 23 & 0 \\
workarena:\textsc{no-submit} & 39 & 0 & 0 & 0 & 0 & 39 & 0 & 0 \\
workarena:\textsc{omit-divergent} & 39 & 23 & 0 & 0 & 0 & 16 & 0 & 0 \\
workarena:\textsc{omit-field} & 13 & 2 & 0 & 9 & 0 & 2 & 9 & 0 \\
\midrule \multicolumn{9}{l}{\emph{Paired experiment of \textsc{Extra-Field} (unit: correct and edited run pair)}} \\
 & Sched. & Integ. & Stop. &  &  & Valid & Pass & FP \\
workarena:\textsc{extra-field}$^{\dagger}$ & 84 & 27 & 21 &  &  & 36 & 36 & 36 \\
\bottomrule
\end{tabular}
\par\vspace{2pt}\begin{minipage}{\linewidth}\footnotesize\raggedright
Upper panel: \texttt{analysis/intent\_swap/fault\_arm\_bounds\_v3.json}. Lower panel: \texttt{analysis/camera\_ready\_2026\_10/outputs/v3\_retally.json}. \textsc{Extra-NI} has no row of its own: of its 48 scheduled runs, 33 were refused and 15 completed with effect evidence; 6 returned \textsc{Pass}. Those 15 are exact re-scores of \textsc{Extra} edited runs, so they add no distinct construction or false positive.
\end{minipage}
\end{table}

In AppWorld, the verifiers rejected every scored run of the four missing-or-wrong-work edits: \textsc{Claim-Only} 48 of 48, \textsc{Omit} 48 of 48, \textsc{Partial} 39 of 39 with 9 refused, and \textsc{Wrong-Value} 33 of 33 with 15 refused. In WorkArena, \textsc{Claim-Only} (27 runs) and \textsc{No-Submit} (39) write no record by construction, and \textsc{Omit-Divergent} (16) omits a task field whose form default differs from the requested value. The verifiers rejected all 82. \textsc{Omit-Field} drops a task field whatever its default. The verifiers passed 9 of its 11 scored runs, and all 9 are harmless edits: the omitted field's default is the requested value, so the record is correct. Because the retained evidence differs across these edits, we count each \textsc{Fail} only as a rejection. In the two templates with false positives, the tests of \texttt{22cc237} build a dictionary from the receiver and amount of each added request (\texttt{dict\_of}) and compare it with the expected shares, and the tests of \texttt{396c5a2} compare \texttt{queue\_song\_ids} by set union.

\label{app:eligibility}
Table~\ref{tab:eligibility} gives the replay's result for each of the 16 audited AppWorld task templates; it classified every task. Eleven templates have only idempotent repeats. Five are eligible, and their verifiers accepted the repeat in two.

\begin{table}[h]
\centering
\caption{Duplicate-write eligibility of the 16 audited AppWorld task templates (48 tasks), grouped by official split. The idempotence replay repeated one write in each task and compared per-table write counters with those of the correct run's replay; it called no verifier. ``Repeated write'' is the endpoint path, with record ids replaced by \{id\}. \textsc{Extra} gives the shipped verifiers' \textsc{Pass} count over the three versions. \textsc{Extra-NI} reproduced every verdict on the five eligible templates. Generated by \texttt{analysis/camera\_ready\_body\_2026\_10/eligibility\_table.py}.}
\label{tab:eligibility}
\footnotesize
\setlength{\tabcolsep}{3pt}
\begin{tabular}{@{}l l l l r l@{}}
\toprule
Template & Split & Repeated write & Replay & \textsc{Extra} & Outcome \\
\midrule
\texttt{0d8a4ee} & dev & \texttt{/phone/messages/text/\{id\}} & adds writes & 0/3 & rejected \\
\texttt{37a8675} & dev & \texttt{/venmo/transactions} & adds writes & 0/3 & rejected \\
\texttt{396c5a2} & dev & \texttt{/spotify/music\_player/song\_queue} & adds writes & 3/3 & false positive \\
\texttt{57c3486} & dev & \texttt{/spotify/songs/\{id\}/like} & no change & 3/3 & harmless edit \\
\texttt{68ee2c9} & dev & \texttt{/file\_system/file/move} & no change & 3/3 & harmless edit \\
\texttt{6c2c621} & dev & \texttt{/file\_system/file} & no change & 3/3 & harmless edit \\
\texttt{d4e9306} & dev & \texttt{/spotify/following\_artists/\{id\}} & no change & 3/3 & harmless edit \\
\texttt{df61dc5} & dev & \texttt{/venmo/transactions/\{id\}/likes} & no change & 3/3 & harmless edit \\
\texttt{22cc237} & train & \texttt{/venmo/payment\_requests} & adds writes & 3/3 & false positive \\
\texttt{2a163ab} & train & \texttt{/venmo/transactions/\{id\}/likes} & no change & 3/3 & harmless edit \\
\texttt{302c169} & train & \texttt{/phone/alarms/\{id\}} & no change & 3/3 & harmless edit \\
\texttt{34d9492} & train & \texttt{/file\_system/file/move} & no change & 3/3 & harmless edit \\
\texttt{7d7fbf6} & train & \texttt{/file\_system/directory/compress} & adds writes & 0/3 & rejected \\
\texttt{aa8502b} & train & \texttt{/spotify/following\_artists/\{id\}} & no change & 3/3 & harmless edit \\
\texttt{b0a8eae} & train & \texttt{/spotify/music\_player/play} & no change & 3/3 & harmless edit \\
\texttt{cf6abd2} & train & \texttt{/simple\_note/notes/\{id\}} & no change & 3/3 & harmless edit \\
\bottomrule
\end{tabular}
\end{table}

Table~\ref{tab:faults} lists the edits whose wrong effect has independent support, from a retained end-state check or from a construction that writes nothing. The paired WorkArena experiment has its own panel.

\begin{table}[t]
\centering
\small
\caption{Edits whose wrong effect has independent support. The Evidence column names either a retained end-state check of the effect or a construction that writes nothing (runner no-write). FP counts false positives. In the WorkArena row ($\dagger$), Runs counts only the complete valid pairs of a correct run and an edited run in the paired experiment, so 36 of 36 is not a rate for the edit. The paired experiment drew its 28 candidate fields from a field scan that never called the verifier, and we committed the manifest before the first paired-experiment episode. The 36 pairs cover 12 candidate fields (seven field names) in the incident and change-request templates, three seeds each. All 57 valid correct runs passed. Of 84 scheduled pairs, 48 were unresolved: in 27 hardware-asset pairs the verifier's configuration failed the frozen configuration check, because WorkArena's constructor writes a per-object serial number into it, and 21 edited runs stopped at a hidden, read-only, or disabled input before the verifier ran. Rows come from different designs, so we do not add their counts.}
\label{tab:faults}
\begin{tabular}{lllrr}
\toprule
Benchmark:edit & Wrong effect & Evidence & Runs & FP \\
\midrule
appworld:\textsc{claim-only} & claim w/o write & runner no-write & 48 & 0 \\
appworld:\textsc{extra} & duplicate write (changed state) & replay write counters & 15 & 6 \\
\midrule workarena:\textsc{claim-only} & claim w/o request & runner no-write & 27 & 0 \\
workarena:\textsc{no-submit} & leave form unsubmitted & runner no-write & 39 & 0 \\
\midrule \multicolumn{5}{l}{\footnotesize\emph{Paired experiment; not comparable with the rows above}} \\
workarena:\textsc{extra-field}$^{\dagger}$ & out-of-scope write & paired Table API check & 36 & 36 \\
\bottomrule
\end{tabular}
\par\vspace{2pt}\begin{minipage}{\linewidth}\footnotesize\raggedright
Table~\ref{tab:fault-flow} gives every verdict of the other registered edits.
\end{minipage}
\end{table}

\section{Reproducibility details}
\label{app:repro}

\paragraph{Audit sets.} On 16 August 2026, before any run, we froze a partition of 34 of AppWorld's train and dev templates into an \emph{audit set} of 20 and a \emph{sealed holdout} of 14 that no run in this paper opened. The 16 state-changing templates of the audit set give 48 tasks, three versions of each. In WorkArena we used all 13 state-changing atomic (L1) templates that the pool's fair-use terms permit, 4 forms and 9 catalog items, at configuration indices 0--2.

\paragraph{Harnesses.} In AppWorld, the harness replays API call programs recorded from the shipped solution on a reset database and compares per-table write counters with those of the task's correct run. In WorkArena, it runs the shipped task driver in a fresh ServiceNow episode with task teardown and a residue check, and the paired experiment reads the saved record through the Table API, which the verifier also queries. Each harness also declares the set $N$ and which actions write, commit, or only claim success. The code for these roles sits in two runners and is not yet one shared interface.

\paragraph{Software pins.} AppWorld: package \nolinkurl{appworld==0.1.3.post1}, data version \texttt{0.1.0}, source commit \nolinkurl{66ad8099e12188ece0d3fe45e661dbc01880813b}, wheel SHA-256 \nolinkurl{db77f8003982502383a50fa2974983894bd1c54f64e2fd3f7e1540d5edd037eb}. WorkArena: package \nolinkurl{browsergym-workarena==0.5.3}, source commit \nolinkurl{a772230a94cf1caf4166b8ead3983f3b3786455b}, wheel SHA-256 \nolinkurl{eb0bdb58598f5e8aeb8b9f01ac3a68dc26d477e3067f01bb56f640a1016e81a5}, with Playwright 1.44.0 and Chromium 125.0.6422.26 (build 1117).

\paragraph{Execution paths.} AppWorld calls the shipped verifier on a fresh execution database for each run. Each edit changes the correct run's API-call program before execution, and the repair regrade runs patched copies of the tests for the six tasks (\S\ref{sec:repairs}). WorkArena runs on a ServiceNow pool instance on the Washington DC release. The runner uses the shipped task driver's \texttt{cheat()} path through BrowserGym and Playwright. When in-session delivery fails, a wrong-task run falls back to a second episode for the trajectory and keeps the verifier's episode open while that episode runs. For \textsc{Extra-Field}, the paired experiment builds the driver from a deep copy of the fixed configuration, adds the candidate to a separate \texttt{fields\_to\_fill} list, and calls the pinned \texttt{\_fill\_fields} routine, which enters values through Playwright label locators before the driver clicks insert. The verifier's \texttt{task\_fields}, \texttt{protected\_fields}, configuration, template record, and goal keep the same hashes and object identities from reset through validation. No run injects a value with JavaScript or writes one through the database API. Neither harness is small: at commit \texttt{2101812e}, the WorkArena grid and paired-experiment runners have 2{,}642 and 1{,}335 lines, and the AppWorld fault, replay, and twin modules have 1{,}340, 931, and 5{,}177. All runs were sequential, because the shared pool instance permits one episode at a time. Wrong-task-grid diagonal runs pass 31 of 31 for AppWorld and 39 of 39 for WorkArena. The AppWorld fault runs also keep 48 of 48 correct runs passing; the WorkArena fault runs reuse the 39 grid correct runs, and the paired experiment runs its own paired correct runs.

\paragraph{Isolation and reset boundary.} AppWorld resets the execution database for each run. WorkArena creates a user for each episode, invokes task teardown, and checks for residue left by that user. Other users can write to the shared instance, so this reset covers only what our episode's user wrote. Interruptions are recorded per run.

\paragraph{WorkArena run context.} The wrong-task grid ran on 23--24 August 2026 against a ServiceNow instance on the Washington DC release (build \nolinkurl{glide-washingtondc-12-20-2023_01-22-2024_1800.zip}). The viewport was 1280$\times$720, \texttt{slow\_mo} was 1{,}000\,ms, the task timeout was 10\,s, and the per-run limit was 420\,s. Every grid reset used seed 7001 with a fixed configuration index; the per-episode Faker user name was deliberately unseeded. The paired experiment ran on 29 August 2026 (UTC) with seeds 20260829, 20260830, and 20260831, on the pool instance with the grid's fingerprint; its preflight release check returned the Washington DC release. The grid artifacts did not retain the host OS or the Python patch version, which leaves a gap for replication.

\section{Artifact availability}
\label{app:artifacts}

A supplementary archive accompanies this paper on OpenReview. Its standard-library reproducer, \texttt{python reproduce.py -{}-check -{}-verify-manifest}, needs Python 3.10 or later and no credentials, model, benchmark server, or verifier. It recomputes the reported counts from the packaged evidence and compares them with a committed expected-results file. The archive holds the frozen protocols and manifests, package pins, the study-stage chronology, the anonymous per-run derivatives from the reviewed submission, the per-family verdict vectors, and the AppWorld rank-sensitivity inputs. It adds aggregate summaries of the camera-ready evidence: the edit tally; the WorkArena paired experiment; the object-identity probe behind the withdrawn WorkArena rerun (Appendix~\ref{app:evidence-correction}); and three model-free AppWorld analyses, which cover repeated create calls in the released baseline runs, reward equality for the six duplicate-write runs, and the count check on natural agent runs; and the per-episode results of the WorkArena per-step experiment (\S\ref{sec:perstep}). The archive contains no AppWorld protected content. AppWorld distributes its task data and evaluators in an encrypted bundle and requires encryption for any public redistribution of them or of their derivatives. The two copied tests of \S\ref{sec:repairs} therefore appear only as SHA-256 digests and a description of the one assertion each inserts, and their twelve regrade ledgers appear only as requirement counts. Hosted ServiceNow credentials and instance identifiers are also excluded. A checksum manifest binds every packaged file. The ZIP is 69{,}532 bytes; its SHA-256 is \nolinkurl{0eac6421c3b1c24c9127394f40038355823b0d1aa7490de4b1ba4faa8643730a}. It replaces the supplement submitted for review. AppWorld and WorkArena are open research benchmarks \citep{trivedi2024appworld,drouin2024workarena}.

\section{Rank orders without an assumed budget}
\label{sec:rank-slice}
\label{app:break-even}

This appendix replaces the reviewed rank-sensitivity table, which assumed a false-positive budget, with break-even rates. The reviewed version asked which of AppWorld's shipped test-split rank orders survive an assumed false-positive budget of 0.4. That budget came from counts over fixed run sets that support no rate, so this appendix assumes no budget. AppWorld's released 2024 baseline archive holds 14 configurations per test split. Take two configurations with shipped Task Goal Completion $q_i > q_j$, a false-positive rate $\alpha$ (the share of truly failed tasks that the verifier passes), and no false rejects. The sharp identification intervals keep $i$ above $j$ exactly when
\[
\alpha \;<\; \alpha^{*}_{ij} \;=\; \frac{q_i - q_j}{1 - q_j}.
\]
On \texttt{test\_normal}, the shipped leader, ReAct with GPT-4o, stays above the runner-up, PlanExec with GPT-4o, for any $\alpha$ below 0.075. Over all 91 pairs, $\alpha^{*}$ runs from 0.009 to 0.479, with median 0.184. On \texttt{test\_challenge}, the same two configurations break even at 0.131, and the 91 pairs run from 0.005 to 0.297, with median 0.097. A positive false-reject rate lowers every $\alpha^{*}$. Recomputed from these values, the reviewed budget of 0.4 leaves 6 of 91 orders on \texttt{test\_normal} and none on \texttt{test\_challenge}, as the reviewed table reported.

Each break-even value is the false-positive rate at which an order would change; it does not show that any order changed. Our audit gives no $\alpha$ for the test splits to compare with them: neither template with false positives is in a test split, and none of the 1{,}277 passing baseline runs repeats the duplicated call (\S\ref{sec:reward}).

\section{Procedure, roles, and steps by experiment}
\label{app:roles}
\label{app:procedure-full}
\label{sec:procedure}

\paragraph{Procedure.} (1)~Freeze the protocol, the enumeration code, and the verifier pins in a commit before the first scored run. (2)~Replay each correct run; drop and report any task whose correct run fails. (3)~Without calling the verifier, apply every edit and record each refusal (the edit cannot apply to that task) with its reason. Run the eligibility scans (idempotence by replay for \textsc{Extra-NI}; form defaults and writable fields for \textsc{Omit-Divergent} and \textsc{Extra-Field}), drop edits that a declared rule marks harmless, and freeze the run manifest. (4)~Deliver each edited run from a fresh start state, paired with its own correct run where the design allows. (5)~Where implemented, check that what the verifier expects hashes the same at every stage, that the driver and the verifier share no mutable object, and that the manifest matches its frozen hash. (6)~Run the end-state check, then (7)~call the verifier. (8)~Give each run exactly one class: refused, delivery incomplete, invalid (a failed correct run or an integrity breach), harmless edit, rejected, unconfirmed candidate, or false positive. (9)~Report the full flow for every edit; pool no edits, and give no rate without a sampling frame. (10)~To test a repair, patch a copy of the verifier by hand and regrade the same runs, or rerun them when the check runs inside the page: the edited run must \textsc{Fail}, and its correct run must still \textsc{Pass}.

\paragraph{Roles.} Table~\ref{tab:human} separates the work a person did from the steps that ran with no per-run human decision and no model call. Not every step of the procedure ran in every experiment. The integrity checks of step~5 ran only in the WorkArena \textsc{Extra-Field} paired experiment, where 57 of 57 valid correct runs, each in its own episode, passed, and in the per-step experiment, which reuses its episode code. Pairing ran there and in the 15 AppWorld \textsc{Extra-NI} re-scores, each with its own correct run. The AppWorld edit runs had one correct run per task (48 of 48 passed) and a write-count end-state check. The wrong-task grids had diagonal runs (31 of 31 and 39 of 39 passed) and no end-state check. The WorkArena missing-or-wrong-work runs reused the grid's correct runs; their \textsc{Claim-Only} and \textsc{No-Submit} runs write nothing by construction.

\begin{table}[h]
\centering
\caption{Who does what in our audit, with the scale of each step in this paper.}
\label{tab:human}
\footnotesize
\begin{tabular}{@{}p{6.45cm}@{\hspace{5mm}}p{6.45cm}@{}}
\toprule
No per-run human decision and no model call & Written or decided by a person \\
\midrule
\textbullet~Enumerating edited runs and recording refusals\newline
\textbullet~Eligibility scans: idempotence, form defaults, writable fields\newline
\textbullet~Distance levels, from the tables each task writes\newline
\textbullet~Delivery, verifier calls, and classification: 2{,}762 wrong-task runs, 459 scheduled missing-or-wrong-work and unrequested-change runs, 15 \textsc{Extra-NI} re-scores (33 of 48 refused before delivery), 168 WorkArena paired-experiment episodes, 84 episodes of the per-step experiment\newline
\textbullet~End-state check: write counters in the AppWorld edit runs, the Table API in the paired experiment and in the per-step experiment; none in the wrong-task grids\newline
\textbullet~The repair tests: the AppWorld regrade, 12 runs; the WorkArena flag-keeping copy, 24 of the 84 episodes of the per-step experiment
&
\textbullet~The edits, with targeting constants read from source\newline
\textbullet~Both harnesses, with their end-state check and the set $N$\newline
\textbullet~The rule that builds AppWorld one-parameter twins\newline
\textbullet~Adding the idempotence replay after the first \textsc{Extra} run\newline
\textbullet~Dropping 8 WorkArena paired-experiment candidates after a stopped first attempt (Appendix~\ref{app:evidence-correction})\newline
\textbullet~The repair patches\newline
\textbullet~Finding the shared-configuration defect in the earlier WorkArena rerun, by reading the code later; no pipeline step flagged it
\tabularnewline
\bottomrule
\end{tabular}
\end{table}

\section{The six preconditions}
\label{app:preconditions}

We claim the method only for a benchmark that meets six preconditions. (P1)~Resettable start states and isolated episodes, with per-episode random values that a correct run and its edited run can share; WorkArena's hardware-asset form gives each task object its own serial number, which broke 27 pairs. (P2)~A correct run that the verifier passes, shipped with the benchmark or produced by a solver. (P3)~Programmatic access to the agent's action channel, so that a run can be edited as a sequence of actions. (P4)~A deterministic verifier that can be called unmodified and regraded as a patched copy; parts that call an LLM judge are excluded or reported apart. (P5)~An end-state check, independent of the verifier's assertions, that sees the effect of every edit, and a declared set $N$. (P6)~A way to pin or hash what the verifier expects. AppWorld had package and wheel pins only; the paired experiment hashed the expected fields and configuration at every stage.

\section{Automated proposers, confirmation and repair}
\label{app:proposer}

A language model could replace the hand-written edits while confirmation stays fixed. It may read the verifier's source and propose candidate loopholes, or act only through the benchmark's normal actions without reading the verifier; either way it confirms nothing itself. A proposed candidate is only a hypothesis. The steps that remain are the ones this paper does by hand: run each candidate from a reset start state, read the end state independently of the verifier to decide whether it is unauthorized, and then patch the verifier so that it rejects the confirmed run without rejecting valid alternative runs, as the natural agent runs of \S\ref{sec:repairs} test for the count assertion. Measuring a proposer's recall would need loopholes planted and sealed by hash before the first run; one execution budget shared by the proposer and its baselines, among them the hand-written edits, random edits, and a model that reads the verifier and lists its loopholes in one call; and, as negative controls, the three AppWorld task templates whose tests already reject the repeat. That study is in progress and reports nothing in this paper.

\end{document}